\documentclass[runningheads]{llncs}
\usepackage{subfigure,lineno,hyperref,graphicx,amsmath,float,multirow,array,booktabs,caption,adjustbox}
\usepackage[T1]{fontenc}
\usepackage{amsmath,amsfonts,bm}

\def\eqref#1{equation~\ref{#1}}
\def\1{\bm{1}}

\def\mP{{\bm{P}}}

\def\mS{{\bm{S}}}

\def\mZ{{\bm{Z}}}

\DeclareMathAlphabet{\mathsfit}{\encodingdefault}{\sfdefault}{m}{sl}
\SetMathAlphabet{\mathsfit}{bold}{\encodingdefault}{\sfdefault}{bx}{n}

\usepackage{graphicx}
\usepackage[font=small,skip=0pt]{caption}
\usepackage{graphicx,adjustbox}
\usepackage[linesnumbered,ruled,vlined]{algorithm2e}
\SetKwComment{Comment}{/* }{ */}
\SetKwComment{LineComment}{\# }{\#}
\SetAlFnt{\small}
\SetKwInput{kwInput}{Input}
\SetKwInput{kwOutput}{Output}

\usepackage[misc]{ifsym}     

\begin{document}
\title{SHIP: A Shapelet-based Approach for Interpretable Patient-Ventilator Asynchrony Detection}
\titlerunning{SHIP: Shapelet-based Interpretable PVA Detection}
%
%\titlerunning{Abbreviated paper title}
% If the paper title is too long for the running head, you can set
% an abbreviated paper title here
%
\author{Xuan-May Le (\Letter)\inst{1} \and
Ling Luo \inst{1}\and
Uwe Aickelin\inst{1} \and
Minh-Tuan Tran\inst{2} \and
David Berlowitz\inst{1} \and
Mark Howard\inst{3}
}
\authorrunning{Xuan-May Le et al.}
% % First names are abbreviated in the running head.
% % If there are more than two authors, 'et al.' is used.
% %
\institute{The University of Melbourne, Melbourne, Victoria, Australia 3010 \\
\email{xuanmay.le@student.unimelb.edu.au} \\ \email{\{ling.luo, uwe.aickelin, david.berlowitz\}@unimelb.edu.au}\\ \and
Monash University, Melbourne, Victoria, Australia 3051 \\
\email{tuan.tran7@monash.edu} \and
Austin Health, Melbourne, Victoria, Australia \\
\email{mark.howard@austin.org.au}
}

\maketitle              % typeset the header of the contribution
\begin{abstract}
Patient-ventilator asynchrony (PVA) is a common and critical issue during mechanical ventilation, affecting up to 85\% of patients. PVA can result in clinical complications such as discomfort, sleep disruption, and potentially more severe conditions like ventilator-induced lung injury and diaphragm dysfunction. Traditional PVA management, which relies on manual adjustments by healthcare providers, is often inadequate due to delays and errors. While various computational methods, including rule-based, statistical, and deep learning approaches, have been developed to detect PVA events, they face challenges related to dataset imbalances and lack of interpretability. In this work, we propose a shapelet-based approach SHIP for PVA detection, utilizing shapelets — discriminative subsequences in time-series data — to enhance detection accuracy and interpretability. Our method addresses dataset imbalances through shapelet-based data augmentation and constructs a shapelet pool to transform the dataset for more effective classification. The combined shapelet and statistical features are then used in a classifier to identify PVA events. Experimental results on medical datasets show that SHIP significantly improves PVA detection while providing interpretable insights into model decisions.

\keywords{Patient-ventilator asynchrony  \and Time Series \and Shapelet.}
\end{abstract}
\section{Introduction}
Patient-ventilator asynchrony (PVA) refers to a mismatch between the timing of the ventilator cycles and the patient's breathing efforts during mechanical ventilation, resulting in a lack of synchronization \cite{pva-definition}. This misalignment is a common issue, affecting up to 85\% of mechanically ventilated patients, and is associated with significant clinical challenges. PVA can negatively impact various aspects of patient health, including disrupting sleep patterns, increasing discomfort, and contributing to respiratory distress \cite{pva-definition,pva-impact-1}. Moreover, prolonged exposure to PVA can lead to more serious complications such as ventilator-induced lung injury and diaphragm dysfunction, ultimately worsening clinical outcomes \cite{pva-impact-2}. We form PVA detection as a multivariate time series classification task, analyzing channels like \textit{Pmask}, \textit{Flow}, \textit{Thor}, and \textit{Abdo} to identify PVA events, such as Autocycling (AC), Double Triggering (DT), and Ineffective Efforts (IE)  \cite{dataset}.

Traditionally, managing PVA has relied on clinical expertise, with healthcare providers manually adjusting ventilator settings to minimize asynchrony. However, this approach is prone to human errors and delays in response, and may not always effectively address the underlying issues \cite{traditional-management-issue-1,traditional-management-issue-2}. To overcome these challenges, various methods have been developed \cite{rule-based-1,statistical-based,spectral-based}. Among these, deep learning techniques \cite{pva-rnn,pva-cnn-1} have shown significant promise in detecting PVA. Despite their potential, these models often face challenges related to \textbf{class imbalances} in training datasets, and their accuracy is closely tied to the specific PVA definitions used. Although these models generally achieve strong performance, their results can vary considerably depending on the network architecture, and they often \textbf{lack interpretability} due to their black-box nature. The lack of transparency presents a major challenge in decision-making, highlighting the need for more interpretable and reliable approaches.

On the other hand, shapelets, class-specific time series subsequences, have proven highly effective in univariate and multivariate time series classification \cite{ppsn,shapenet,shapeformer,shapelet}. Their success lies in their ability to capture class-specific information, as the distance between a shapelet and time series from the same class is significantly smaller than for other classes. Shapelets are intuitive and discriminative, making them powerful tools for identifying local patterns that distinguish between different classes, which is crucial for time series classification tasks.

\begin{figure}[t]
\begin{center}
\includegraphics[width=0.65\linewidth]{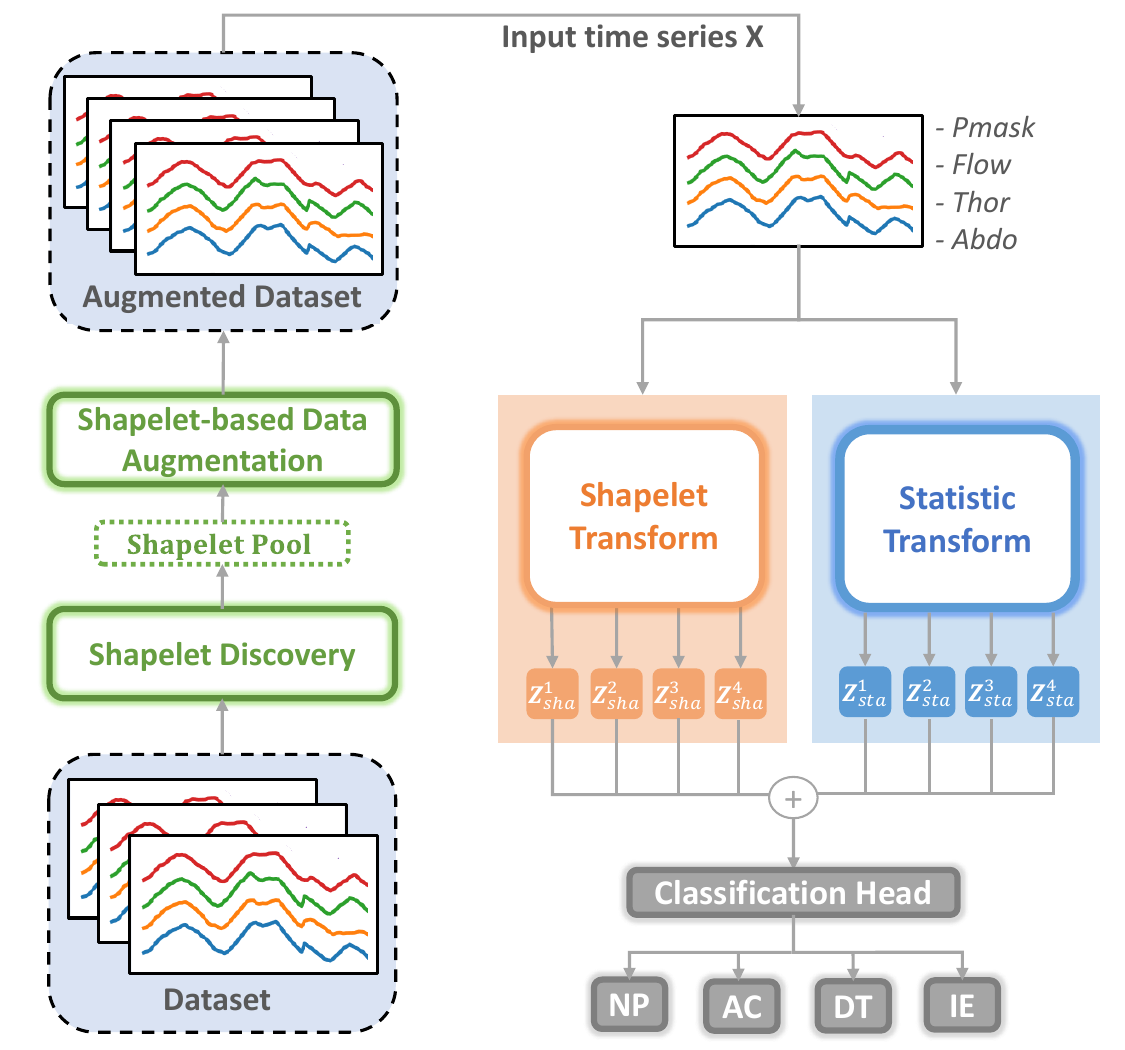}
\end{center}
\caption{The general architecture of our proposed method for detecting PVA events (Autocycling (AC), Double Triggering (DT), and Ineffective Efforts (IE)) and Non-PVA (NP).}
\label{fig:pva}
\end{figure}

In this work, we propose a \textbf{SH}apelet-based Approach for \textbf{I}nterpretable \textbf{P}atient-Ventilator Asynchrony Detection (SHIP) to address the aforementioned problems. The overall architecture of our proposed method is illustrated in Figure \ref{fig:pva}. First, we apply shapelet-based data augmentation approach to enhance rare events, addressing the issue of imbalanced data. Next, we extract shapelets from the augmented dataset, selecting high-quality shapelets to form a shapelet pool. The augmented dataset is then transformed using these shapelets. Finally, the shapelet features are combined with statistical features and fed into a classifier to accurately identify PVA events. By leveraging shapelets for this task, we can effectively tackle imbalanced data while maintaining interpretability, as indicated in Figure \ref{fig:inter}. The interpretability of shapelets for PVA detection is evident. Shapelets belonging to a specific class tend to exhibit a closer distance to instances within that class. Consequently, given a time series, our model's decision can be interpreted by analyzing the distance between the shapelets of the predicted class and the time series.

Our contributions can be summarized as follows: (i) We propose a novel and interpretable shapelet-based approach, SHIP, to detect PVA events; (ii) We introduce shapelet-based data augmentation to address the issue of dataset imbalance; (iii) Experimental results on medical datasets demonstrate that SHIP improves PVA detection while providing interpretable insights into the model's decisions; (iv) Notably, our method performs well using only two channels, \textit{Pmask} and \textit{Flow}, achieving results comparable to those obtained with four channels.

\section{Related Work}

Various methods have been developed to address PVA detection, including rule-based approaches \cite{rule-based-1,rule-based-2} that use pre-defined criteria to detect asynchronies; model-based techniques \cite{model-based-2} relying on physiological or mechanical models; statistical methods \cite{statistical-based} using probabilistic analyses; spectral approaches \cite{spectral-based} that analyze frequency-domain characteristics; and machine learning-based approaches \cite{machine-learning-based-1,machine-learning-based-2,knn-1,knn-2} that utilize the classic machine learning model for this task. However, these methods often suffer from low performance due to challenges in capturing the complexity and variability of the PVA's time series data. 

In recent years, deep learning methods have demonstrated potential in PVA detection. For instance, \cite{pva-rnn} leveraged rule-based PVA definitions to train recurrent neural network (RNN) models, whereas \cite{pva-cnn-1,pva-cnn-2} utilized convolutional neural networks (CNNs) to identify asynchronies from PVA waveforms. Additionally, \cite{gc} introduced a conditional latent Gaussian mixture generative classifier for this purpose. Although these models show potential, they often struggle with class imbalances in training datasets and depend on specific PVA definitions. Their performance can vary widely depending on network architecture, and they lack interpretability due to their black-box nature, posing a significant challenge for more robust and transparent solutions.
    . 

\section{Methodology}
\subsection{PVA Dataset}

The data for this research was collected as part of a randomized controlled trial approved by the Research Ethics Board of Austin Health in Melbourne, Australia \cite{dataset}. The study involved 59 patients in two overlapping groups who underwent polysomnographic (PSG) titration while receiving non-invasive ventilation to assess the impact of PSG titration on sleep quality and patient-ventilator asynchrony (PVA). We analyzed ventilation monitoring data from the participants, resulting in 104 raw data samples, each comprising approximately 20 to 24 data channels. Along with standard ventilator metrics such as mask pressure (\textit{Pmask}) and flow (\textit{Flow}), the dataset includes readings from an effort belt device on the thoracic and abdominal regions, referred to as \textit{Thor} and \textit{Abdo}, respectively. These readings approximate breathing effort and are relevant to PVA events. The dataset also includes sleep stage scoring derived from EEG recordings \cite{dataset}. A respiratory expert manually annotated the dataset, identifying PVA and sleep disruption events based on PSG recordings. Although several types of PVA events exist, we focused on Double Triggering (DT), Autocycling (AC), and Ineffective Efforts (IE), which account for over 90\% of all PVA occurrences \cite{pva-events-1,pva-events-2}. Specifically, (i) DT occurs when the ventilator delivers two consecutive breaths with very short expiratory time; (ii) AC is similar to DT but with more than two breaths; (iii) IE occurs when an inspiratory muscle effort is not followed by a ventilator breath.

\subsection{Data Preprocessing} 

Inspired by \cite{gc}, we focused on four key channels: \textit{Pmask}, \textit{Flow}, \textit{Thor}, and \textit{Abdo}, selected for their relevance in detecting PVA. First, we preprocessed the dataset to ensure it was clean and suitable for analysis. We then segmented the raw data based on the mask pressure ($P_\text{mask}$) channel and applied zero-padding to standardize sample lengths to 150, corresponding to the maximum length of a single breathing cycle. After segmentation, we obtained 305,130 time series instances, categorized into four groups: NP (non-PVA events), AC, DT and IE. The distribution was 280,110 NP instances, 6,385 AC, 10,595 DT, and 8,040 IE. This underscores the presence of a notable class imbalance, with the non-PVA category having a considerably higher number of instances compared to the PVA-related events. For model training and evaluation, 80\% of the dataset was randomly allocated for training, and 20\% for validation.

\noindent
\textbf{PVA Detection Task.} We represent the multivariate time series as $\mathbf{X} \in \mathbb{R}^{V \times T}$, where $V = 4$ denotes the number of channels and $T = 150$ represents the length of the time series. Here, $\mathbf{X} = \{x^{1,1}, \ldots, x^{V,T}\}$, where $x^{v,t}$ signifies the value for channel $v$ at timestamp $t$ within $\mathbf{X}$. Consider a training dataset $\mathcal{D} = \{(\mathbf{X}_i, y_i)\}_{i=1}^M$, where $M$ is the number of time series instances, $\mathbf{X}_i$ represents a training instance, and $y_i$ is its corresponding label ($y_i \in \mathcal{Y}$). The objective of PVA detection is to train a classifier $f(\mathbf{X})$ to predict a class label $y$ for $\mathbf{X}$.

\subsection{Shapelet Discovery}
\label{sec:se}

\begin{algorithm}[t]
\caption{Offline Shapelet Discovery}
\label{alg:opse}
\kwInput{$\mathcal{D}$: dataset; time series length $T$; channels $V$; number of PIPs $k$; number of shapelets $g$; classes $\mathcal{Y}$ with $|Y|$ as the number of classes.}

% \Comment{Shapelet Extraction}
$\mathcal{C} = []$, $\mathcal{S} = []$ \LineComment{Candidate and Shapelet sets;}
\ForEach{$(\mathbf{X}, y) \in \mathcal{D}$}{
    \For{$v = 1$ to $V$}{
        $\mP = [1, T]$ \LineComment{PIPs set;}
        \For{$j = 1$ to $k - 2$}{
            Find index $p$ maximizing reconstruction distance\;
            Add $p$ to $\mP$, sorted and get index $idx$ of $p$ in $\mP$\;
            \For{$z = 0$ to $2$}{
                \If{$idx+2-z \leq |\mP|$ and $idx-z \geq 1$}{
                     $p_s = \mP[idx-z], p_e = \mP[idx+2-z]$ \;
                     Define candidate ${C}$ from $\mathbf{X}[p_s:p_e]$\; 
                     Add [${C}$, $p_s$, $p_e$, $v$, $y$] to $\mathcal{C}$\;
                }
            }
        }
    }
}
% \Comment{Shapelet Selection}
\ForEach{$(C, y) \in \mathcal{C}$}{
    Calculate the information gain of $C$ for class $y$ using Eq.~\ref{eq:psd} with all $\mathbf{X}_i \in \mathcal{D}$\;
}

\ForEach{$\tilde{Y} \in \mathcal{Y}$}{
    Select top $g/|\mathcal{Y}|$ candidates $S \in \mathcal{C}$ by information gain and add to $\mathcal{S}$\;
}
\textbf{return} $\mathcal{S}$
\end{algorithm}
This section introduces the Offline Shapelet Discovery (OSD) method, designed to efficiently extract small, high-quality shapelets from time series data by utilizing Perceptually Important Points (PIPs) \cite{pip,pisd}. The selection process is driven by the reconstruction distance, defined as the perpendicular distance between a target point and the line formed by the nearest PIPs \cite{ppsn,pip}. The method consists of two main phases: shapelet extraction and shapelet selection, with the pseudo-code provided in Algorithm \ref{alg:opse}. Firstly, OSD identifies shapelet candidates by selecting PIPs. It begins by adding the first and last indices to the PIPs set, then iteratively includes the index with the highest reconstruction distance. Each time a new PIP is added, up to three shapelet candidates are generated from three consecutive PIPs. For each shapelet, four attributes are stored: the shapelet value vector, start index, end index, and channels. Secondly, OSD selects an equal number of shapelets for each class. For each shapelet candidate \( S_i \) of class \( \tilde{Y} \), we calculate its Perceptual Subsequence Distance \cite{ppsn} with all training instances, as shown in Eq.~\ref{eq:psd}.

\noindent
\textbf{Perceptual Subsequence Distance (PSD)}: Given a time series \( \mathbf{X} \) of length \( T \), and a subsequence \( S = [s_1, \dots, s_l] \) of length \( l \), with \( l \leq T \), the PSD of \( \mathbf{X} \) and \( S \) is determined as:
\begin{equation}
    PSD(\mathbf{X}, S) = \min_{j=1}^{T-l+1} \left( \text{CID}(\mathbf{X}[j:j+l-1], S) \right),
    \label{eq:psd}
\end{equation}
where CID is the complexity-invariant distance, a measure commonly used in time series mining \cite{cid,csax} and specifically in shapelet discovery \cite{ppsn}.

The PSD can guide the selection of candidates with optimal information gain, with the highest achievable gain by shapelet \( S_i \) indicating the top choices \cite{ppsn}. Finally, the top \( g \) shapelets with the highest information gain are selected and stored in the shapelet pool \( \mS \).

% For each class $\tilde{y} \in \mathcal{Y}$ and $\tilde{g} = g/|\mathcal{Y}|$ is the number of shapelet for class $\tilde{y}$, given their set of shapelets $ \mS^{\tilde{y}} = [S^1, \ldots, S^{\tilde{g}}]$ and the subset $\mathcal{D}^{\tilde{y}} = \{(\mathbf{X}_i, y_i)\}_{i=1}^{M^{\tilde{y}}}$ containing all the instances of class $\tilde{y}$. Their augmented data $\mathbf{\tilde{X}}$ is calculated as follows:
\subsection{Shapelet-based Data Augmentation}

In this section, we propose shapelet-based data augmentation for the minority classes (e.g., PVA events such as AC, DT, and IE). For each instance of these classes, we suggest augmenting it to generate $r_{sa} = 10$ additional instances (we also provide the ablation study in Section \ref{sec:aa} to choose this ratio), thereby addressing the data imbalance issue. Given the time series instances $\mathbf{\tilde{X}} = [\tilde{x}^{1,1}, \ldots, \tilde{x}^{V,T}] \in \mathcal{D}$ of class $\tilde{Y}$ and a shapelet $\tilde{S} \in \mathbf{S}$ of class $\tilde{Y} \in \mathcal{Y}$, their augmented data $\mathbf{\hat{X}}= [\hat{x}^{1,1}, \ldots, \hat{x}^{V,T}]$ is calculated in the following steps.
\begin{align}
    \hat{\mathbf{X}} = \tilde{\mathbf{X}} +  \mathbf{\mathcal{E}} \odot \mathbf{\mathcal{M}},
\end{align}
where $\mathbf{\mathcal{E}} = [\epsilon^{1,1}, \dots, \epsilon^{V,T}] \sim \mathcal{N}(\mu, \sigma^2)$ represents Gaussian noise, $\odot$ denotes element-wise multiplication, and the mask $\mathbf{\mathcal{M}} \in \mathbb{R}^{V \times T}$ is calculated as follows:

We first find the best matching subsequences $\mathbf{B}$ of shapelets $\tilde{S}$ in the time series $\mathbf{\tilde{X}}$ with minimal $PSD(\mathbf{B}, \tilde{S})$. Then, the mask $\mathbf{\mathcal{M}} = [m^{1,1}, \ldots, m^{V,T}]$ is calculated as:
\begin{align}
    m^{v,t} = 
    \begin{cases} 
    1 & \text{if } \tilde{x}^{v,t} \notin \mathbf{B}, \\ 
    PSD(\mathbf{B}, \tilde{S}) & \text{if } \tilde{x}^{v,t} \in \mathbf{B}.
    \end{cases}
\end{align}
If $x^{v,t} \notin \mathbf{B}$, noise is added as $\hat{x}^{v,t} = \tilde{x}^{v,t} + \epsilon^{v,t}$. Otherwise, the noise is scaled, and $\hat{x}^{v,t} = \tilde{x}^{v,t} + PSD(\mathbf{B}, S) \cdot \epsilon^{v,t}$. When $PSD(\mathbf{B}, S)$ is small, the value of $\mathbf{X}$ remains nearly unchanged. This process ensures that the augmented instances are different from the original data while still retaining important shape characteristics specific to their class.

\subsection{Shapelet Transform} In this section, our method transforms the time series based on the PSD between a given time series and a set of selected shapelets (Eq. \ref{eq:psd}). The goal of this transformation is to represent each time  series based on its similarity to each shapelet, creating a transformed vector that reflects the presence and strength of shapelet patterns in the original time series. Given the set of shapelets $\mS = [S^1, \ldots, S^g]$ and the input time series $\mathbf{X}$. The transformed vector, $\mathbf{Z}_{\text{sha}} = [Z^1_{\text{sha}}, \ldots, Z^g_{\text{sha}}]$ of $\mathbf{X}$ is computed as follows:
\begin{equation}
    Z^j_{\text{sha}} = PSD(\mathbf{X}, S^j), \quad\forall j \in  [1,.., g]
\end{equation}

This means that for each shapelet $S^j$, we calculate $Z^j_{\text{sha}}$, which is the PSD between $\mathbf{X}$ and $S^j$. This PSD value quantifies how closely $\mathbf{X}$ aligns with the pattern represented by $S^j$, where smaller values indicate a closer match. Repeating this process for all shapelets $S^j \in \mathbf{S}$ yields the transformed vector $\mathbf{Z}_{\text{sha}}$, which encodes how each shapelet is reflected in $\mathbf{X}$. This transformed vector $\mathbf{Z}_{\text{sha}}$ can then be used as input features for subsequent classification or analysis.

\subsection{Statistic Transform}
Building on the approach in \cite{gc}, we derive statistical features from time series data to characterize its temporal dynamics. The statistical feature vector $\mathbf{Z}_{\text{sta}}$ for a given time series $\mathbf{X}$ is computed as follows:
\begin{equation}
\mathbf{Z}_{\text{sta}} = \text{LogSig}(\mathbf{X_i}),
\end{equation}
where $\text{LogSig}(\mathbf{X})$ represents the logarithmic signature of the time series $\mathbf{X}$. The logarithmic signature is a collection of features that capture the temporal structure of the time series, and is mathematically defined as:
\begin{equation}
\text{LogSig}(\mathbf{X}) = \left(\text{LogSig}^1(\mathbf{X}), \ldots, \text{LogSig}^N(\mathbf{X}) \right),
\end{equation}
where $\text{LogSig}^n(\mathbf{X})$ denotes the $n^{\text{th}}$ order term of the logarithmic signature, which is computed as:
\begin{equation}
\text{LogSig}^n(\mathbf{X}) = \sum_{1 \leq i_1 < \ldots < i_n \leq n} \log(x_{i_n} - x_{i_{n-1}}).
\end{equation}
In this formula, $x_i$ denotes the time points in the series, and the sum is performed over all combinations of $n$ indices from the time series, ensuring they are in increasing order. This process aggregates the logarithmic differences between the time points, capturing important information about the temporal dynamics of the series.

\subsection{Final Classification Head}
The final classification head combines transformed features and produces a label prediction. First, we concatenate the shapelet and static feature vectors, $\mathbf{Z}_{\text{sha}}$ and $\mathbf{Z}_{\text{sta}}$, into a single vector, $\mathbf{Z} = \text{concat}(\mathbf{Z}_{\text{sha}}, \mathbf{Z}_{\text{sta}})$. Given $d$ is dimension of $\mZ$, $\mZ$ is then passed through three linear layers to reduce dimensions sequentially: from $d$ to 512, then 512 to 256, and finally 256 to 4 (i.e., the number of classes). After applying softmax to convert the output to a predicted probability distribution.
\begin{align} 
\mathbf{Z} &= \text{Linear}(\text{Linear}(\text{Linear}(\mathbf{Z}))) \\
\hat{y}_i &= \text{softmax}(\mathbf{Z})
\end{align}
The model’s predictions $\hat{y}$ are optimized by minimizing the Cross-Entropy Loss $\mathcal{L}_{\text{CE}}(\hat{y}, Y) = -\sum_{i=1}^{|Y|} y_i \log(\hat{y}_i)$, which encourages the model to assign higher probabilities to the correct classes.
\section{Experiment}
\subsection{Experimental Setting}

In our experiments, we assess the performance of SHIP on PVA events using a medical dataset \cite{dataset}. For hyperparameter tuning, specifically for window size $w$ and the number of PIPs $k$, we employ leave-one-out cross-validation (LOOCV). We evaluate 10 distinct values of $k$, dividing the range evenly from 3 to $0.1 \times T$. All experiments were conducted on a machine with a single NVIDIA A100 40GB GPU.

\noindent
\textbf{Evaluation Metric.} Accuracy and F1-score serve as the primary metrics for comparison. We also report average precision, recall, and per-class F1-score to demonstrate the effectiveness of the methods in addressing PVA detection with data imbalance issue.

\noindent
\textbf{Compared Baselines.} We evaluate the performance of our proposed SHIP model by comparing it against several well-established methods. Specifically, we use a classic Recurrent Neural Network (RNN) \cite{pva-rnn}, a Convolutional Neural Network (CNN) \cite{pva-cnn-1}, and a Gaussian mixture classifier (GC) \cite{gc}, which is one of the commonly employed approaches for machine learning classification tasks. 

%using four channels, \textit{Pmask}, \textit{Flow}, \textit{Thor}, \textit{Adbo}

\subsection{Classification Result}
\begin{table}[t]
\centering
\caption{The performance metrics for four different models : CNN, RNN, GC, and SHIP. The metrics include the F1 scores for each class including Non-PVA (NP), Autocycling (AC), Double Triggering (DT), and Ineffective Efforts (IE), as well as overall Precision, Recall, F1-score, and Accuracy.}
\begin{adjustbox}{width=0.8\linewidth}
\begin{tabular}{lcccccccc}
\hline
              & F1(NP)          & F1(AC)          & F1(DT)          & F1(IE)          & Precision       & Recall          & F1-score        & Accuracy        \\ \hline
CNN           & 0.8050          & 0.2371          & 0.2524          & 0.2050          & 0.6696          & 0.6770          & 0.6698          & 0.6770          \\
RNN           & 0.8117          & 0.2980          & 0.2522          & 0.2030          & 0.6767          & 0.6872          & 0.6786          & 0.6872          \\
GC            & 0.9830          & 0.8752          & 0.7530          & 0.6458          & 0.9645          & 0.9641          & 0.9638          & 0.9641          \\
\textbf{SHIP} & \textbf{0.9893} & \textbf{0.9015} & \textbf{0.8894} & \textbf{0.7050} & \textbf{0.9764} & \textbf{0.9773} & \textbf{0.9765} & \textbf{0.9773} \\ \hline
\end{tabular}
\end{adjustbox}
\label{tab:facc}
\end{table}

Table \ref{tab:facc} shows that SHIP outperforms all other models across all metrics, achieving the highest F1 scores of 0.9893 for NP, 0.9015 for AC, 0.8894 for DT, and 0.7050 for IE, demonstrating its robustness in classifying each class. SHIP also achieved a Precision of 0.9764, Recall of 0.9773, overall F1-score of 0.9765, and Accuracy of 0.9773. While GC performs well overall, it struggles with the IE class (F1-score of 0.6458). In comparison, CNN and RNN models show lower performance, particularly on AC and IE, with F1 scores below 0.3. Overall, these results emphasize the superior classification capabilities of the SHIP model, making it the most effective choice among the evaluated models for this dataset.

\noindent
\textbf{PVA Detection with Two Channels:} Collecting \textit{Thor} and \textit{Abdo} data usually requires specialized equipment, like respiratory inductance plethysmography belts or other sensors that measure chest and abdominal movement. These sensors can be costly and may add extra setup time. Therefore, it is desirable to detect PVA using only \textit{Pmask} and \textit{Flow}. Table \ref{tab:2dim} compares the performance of SHIP with baseline models using two channels. SHIP outperforms most models, achieving F1 scores of 0.9860 for NP, 0.8943 for AC, and 0.8670 for DT, with a lower score of 0.5015 for IE. CNN and RNN perform poorly, especially for AC (F1 scores of 0.2139 and 0.2229). The GC model performs well for NP (F1 of 0.9838) but lags behind SHIP overall. SHIP's high Precision (0.9681), Recall (0.9706), and Accuracy (0.9706) highlight its effective use of two-channel for improved classification. Overall, the results of PVA detection using two channels, \textit{Pmask} and \textit{Flow} are highly satisfactory; however, performance significantly drops for the IE event. Clinically, this indicates that the detection of IE events is primarily associated with the two belt-related channels (i.e., \textit{Thor} and \textit{Abdo}).

\begin{table}[t]
\centering
\caption{Comparison between our SHIP model and the baselines using only two channels (\textit{Pmask} and \textit{Flow}).}
\begin{adjustbox}{width=0.7\linewidth}
\begin{tabular}{lcccccccc}
\hline
              & F1(NP)         & F1(AC)          & F1(DT)          & F1(IE)          & Precision       & Recall          & F1-score        & Accuracy        \\ \hline
CNN           & 0.7484          & 0.2139          & 0.1037          & 0.0821          & 0.5594          & 0.5725          & 0.5653          & 0.5725          \\
RNN           & 0.7626          & 0.2229          & 0.1929          & 0.1764          & 0.5849          & 0.6053          & 0.5930          & 0.6053          \\
GC   & 0.9838                 & 0.7167                 & 0.6860                 & 0.4857                 & 0.9538               & 0.9556               & 0.9543               & 0.9556               \\
\textbf{SHIP} & \textbf{0.9860} & \textbf{0.8943} & \textbf{0.8670} & \textbf{0.5015} & \textbf{0.9681} & \textbf{0.9706} & \textbf{0.9672} & \textbf{0.9706} \\ \hline
\end{tabular}
\end{adjustbox}
\label{tab:2dim}
\end{table}
\begin{table}[t]
\caption{Comparison of the baseline \cite{gc} and our method variants: `S' using shapelet features; `SA' using shapelet-based augmentation; and `SA' (SHIP), combining both.}
\centering
\begin{adjustbox}{width=0.7\linewidth}
\begin{tabular}{lcccccccc}
\hline
\multicolumn{1}{c}{}        & F1(NP)         & F1(AC)          & F1(DT)          & F1(IE)          & Precision       & Recall          & F1-score        & Accuracy        \\ \hline
Baseline                    & 0.9830          & 0.8752          & 0.7530          & 0.6458          & 0.9645          & 0.9641          & 0.9638          & 0.9641          \\
S                   & 0.9871          & 0.8838          & 0.8738          & 0.6610          & 0.9723          & 0.9728          & 0.9724          & 0.9728          \\
SA & 0.9874                 & 0.8908                 & 0.7653                 & 0.6893                 & 0.9715               & 0.9705               & 0.9701               & 0.9705               \\
S+SA & \textbf{0.9893} & \textbf{0.9015} & \textbf{0.8894} & \textbf{0.7050} & \textbf{0.9764} & \textbf{0.9773} & \textbf{0.9765} & \textbf{0.9773} \\ \hline
\end{tabular}
\end{adjustbox}
\label{tab:abs}
\end{table}
\subsection{Ablation Study}
\label{sec:aa}

\noindent
\textbf{Component Evaluation.} The ablation study evaluates the performance of a baseline model \cite{gc} against three variants of our method: `S', using shapelet features; `SA', using shapelet-based augmentation; and `S+SA' (SHIP), combining both. The results, shown in Table \ref{tab:abs}, demonstrate that the `S' variant improves performance, yielding higher F1 scores, especially for NP and DT, with an overall Accuracy of 0.9728. Similarly, the `SA' variant shows competitive results, significantly improving the F1 score for IE to 0.7913 while maintaining an overall Accuracy of 0.9723. However, SHIP outperforms all variants, achieving the highest F1 scores of 0.9893 for NP, 0.9015 for AC, 0.8894 for DT, and 0.7050 for IE, with an impressive overall Accuracy of 0.9773. These results highlight that combining shapelet features with shapelet-based augmentation significantly boosts the performance, underscoring the effectiveness of our proposed method.

\noindent
\textbf{Different Augmentation Ratios $r_{sa}$.} In Table \ref{tab:aa}, we further compare our method with different augmentation ratios. Smaller values result in significantly lower performance for minority classes, while higher values can drastically reduce performance for majority classes. Therefore, the highest performance is achieved at \(r_{sa} = 10\), as it strikes a balance between the minority and majority classes.

\begin{table}[t]
\caption{Comparison of different ratios for shapelet-based augmentation.}
\centering
\begin{adjustbox}{width=0.7\linewidth}
\begin{tabular}{@{}lcccccccc@{}}
\toprule
\multicolumn{1}{c}{$r_{sa}$} & F1(NP)          & F1(AC)          & F1(DT)          & F1(IE)          & Precision       & Recall          & F1-score        & Accuracy        \\ \midrule
5             & 0.9894          & 0.8826          & 0.8422          & 0.6606          & 0.9741          & 0.9749          & 0.9741          & 0.9749          \\
10            & \textbf{0.9893} & \text{0.9015} & \textbf{0.8894} & \text{0.7050} & \textbf{0.9764} & \textbf{0.9773} & \textbf{0.9765} & \textbf{0.9773} \\
20            & 0.9884          & \textbf{0.9053 }         & 0.8722          & \textbf{0.7149}          & 0.9746          & 0.9756          & 0.9748          & 0.9756          \\ \bottomrule
\end{tabular}
\end{adjustbox}
\label{tab:aa}
\end{table}

\subsection{Interpretable Visualization}

%In Fig. \ref{fig:inter}, it is evident that our model's classification decisions primarily rely on the distance between the observed time series and the corresponding shapelet for each different PVA event. This distance metric serves as a key criterion for determining PVA events, enhancing the interpretability of the model's predictions. A shorter distance indicates a stronger similarity to the shapelet, allowing practitioners to understand why a particular event was assigned based on recognizable patterns within the time series. Furthermore, shapelets encapsulate distinctive characteristics of the data, making them inherently interpretable features. By tracing the classification back to the specific patterns represented by the shapelet, clinicians gain insights into the model's reasoning. 

\begin{figure}[t]
\begin{center}
\includegraphics[width=0.7\linewidth]{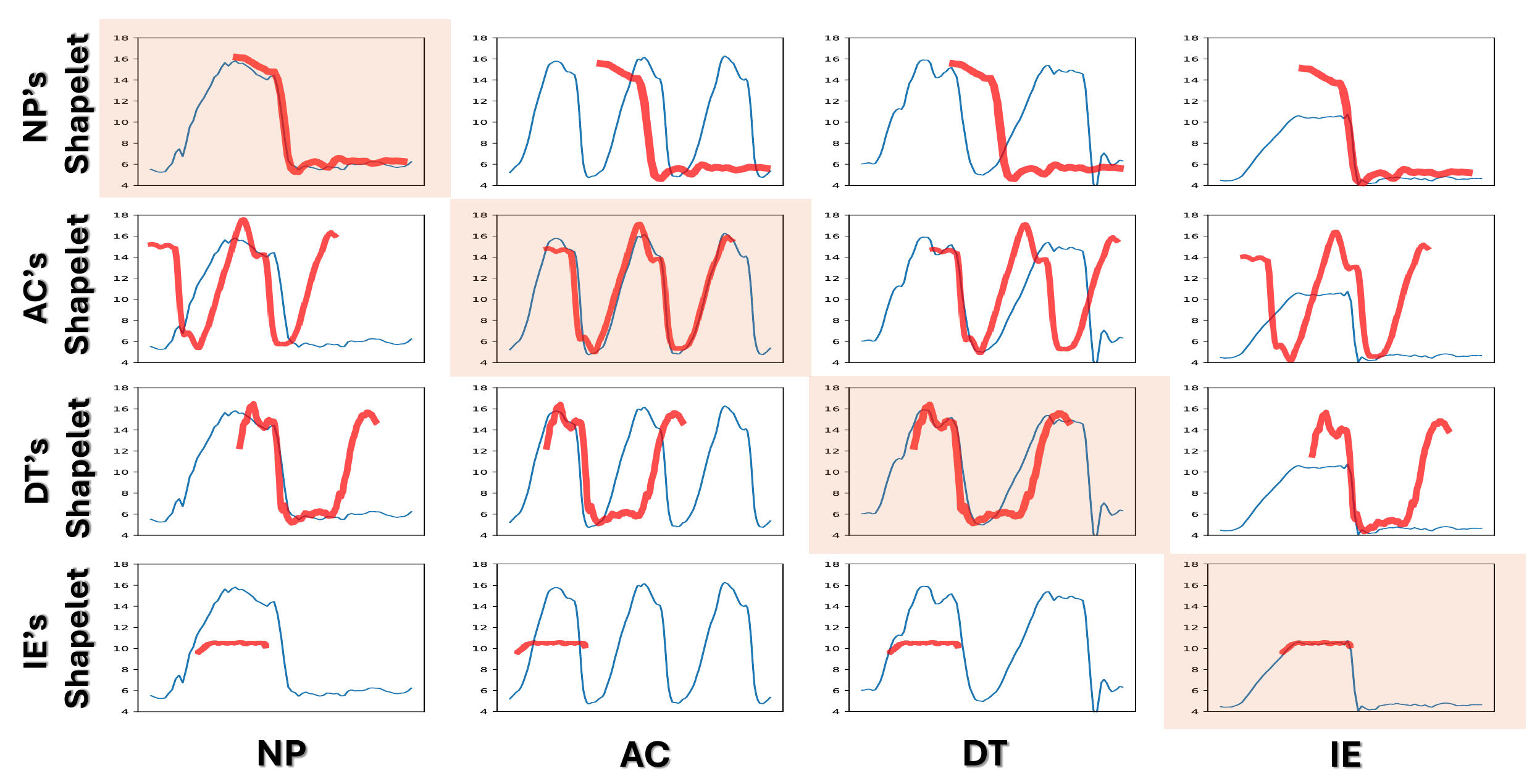}
\end{center}
\caption{The interpretability of shapelets for PVA detection, including four types of events (NP, AC, DT, IE).}
\label{fig:inter}
\end{figure}

In Fig. \ref{fig:inter}, the classification decisions made by our model are primarily driven by the distance between the observed time series and the corresponding shapelet for each PVA event. For instance, the AC shapelet (second row) exhibits a distinct pattern that closely aligns with the observed time series, indicating a high similarity to the AC event. Similarly, the DT shapelet (third row) corresponds to a series of trigger events that match the shapelet, reinforcing the model's classification of this time series as DT. This distance metric plays a crucial role in identifying PVA events, enhancing the interpretability of the model’s predictions. A smaller distance reflects a stronger match to the shapelet, enabling clinicians to understand the rationale behind the classification based on recognizable patterns in the time series. Moreover, shapelets encapsulate distinctive features of the data, making them naturally interpretable. By tracing the model's predictions back to the specific patterns represented by the shapelets, clinicians can gain deeper insights into the model’s decision-making process.

\section{Conclusion}

In conclusion, our SHapelet-based Interpretable Patient-Ventilator Asynchrony Detector (SHIP) addresses key challenges in PVA detection, such as data imbalance and model interpretability. By using shapelet-based data augmentation, SHIP improves accuracy for rare PVA events and enhances transparency, aiding clinicians in understanding model outputs. Experimental results highlight SHIP's reliability in detecting PVA events, making it a valuable tool for improving patient care in mechanical ventilation. Future work will explore integrating SHIP into real-time monitoring systems for dynamic detection and intervention.

\end{document}